\documentclass[manuscript]{acmart}

\AtBeginDocument{%
  \providecommand\BibTeX{{%
    \normalfont B\kern-0.5em{\scshape i\kern-0.25em b}\kern-0.8em\TeX}}}

\acmYear{2024}
\copyrightyear{2024}
\setcopyright{rightsretained}
\acmConference[ACM FAccT '24]{ACM Conference on Fairness, Accountability, and Transparency}{June 3--6, 2024}{Rio de Janeiro, Brazil}
\acmBooktitle{ACM Conference on Fairness, Accountability, and Transparency (ACM FAccT '24), June 3--6, 2024, Rio de Janeiro, Brazil}
\acmDOI{10.1145/3630106.3658542}
\acmISBN{979-8-4007-0450-5/24/06}

\acmSubmissionID{facct2024-6}

\usepackage{multirow}
\usepackage{colortbl}
\usepackage{booktabs}
\usepackage{hwemoji}


\begin{document}


\title[Power Hungry Processing]{Power Hungry Processing: {⚡} Watts {⚡} Driving the Cost of AI Deployment?}

\author{Alexandra Sasha Luccioni}
\email{sasha.luccioni@huggingface.co}
\author{Yacine Jernite}
\affiliation{%
  \institution{Hugging Face}
  \country{Canada/USA}
}

\author{Emma Strubell}
\affiliation{%
  \institution{Carnegie Mellon University, Allen Institute for AI}
  \country{USA}}

\renewcommand{\shortauthors}{Luccioni et al}

\begin{abstract}
Recent years have seen a surge in the popularity of commercial AI products based on generative, multi-purpose AI systems promising a unified approach to building machine learning (ML) models into technology. However, this ambition of ``generality'' comes at a steep cost to the environment, given the amount of energy these systems require and the amount of carbon that they emit. In this work, we propose the first systematic comparison of the ongoing inference cost of various categories of ML systems, covering both task-specific (i.e. finetuned models that carry out a single task) and `general-purpose' models, (i.e. those trained for multiple tasks). We measure deployment cost as the amount of energy and carbon required to perform 1,000 inferences on representative benchmark dataset using these models. We find that multi-purpose, generative architectures are orders of magnitude more expensive than task-specific systems for a variety of tasks, even when controlling for the number of model parameters. We conclude with a discussion around the current trend of deploying multi-purpose generative ML systems, and caution that their utility should be more intentionally weighed against increased costs in terms of energy and emissions. All the data from our study can be accessed via an \href{https://huggingface.co/spaces/sasha/CO2_inference/}{interactive demo} to carry out further exploration and analysis.
\end{abstract}



\begin{teaserfigure}
\centering
\includegraphics[width=0.8\textwidth]{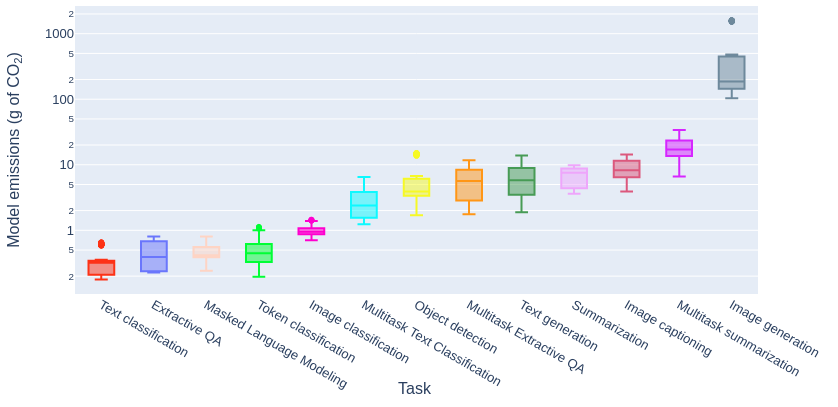}
\caption{The tasks examined in our study and the average quantity of carbon emissions they produced (in g of $CO_2eq$) for 1,000 queries. N.B. The y axis is in logarithmic scale.}
\label{fig:tasks-carbon}
\end{teaserfigure}

\maketitle

\section{Introduction}

Understanding the environmental impacts of different industries is an important first step towards developing effective strategies to mitigate those impacts. For newer industries such as information and communication technologies (ICT) of which Artificial Intelligence (AI) and Machine Learning (ML) are considered to be a part of, more work is needed to understand the extent of their environmental impacts and the factors that influence it. Between 2017 and 2021, the electricity used by Meta, Amazon, Microsoft, and Google, the main providers of commercially-available cloud compute, more than doubled ~\cite{IEA2023}. According to the most recent figures available, global data centre electricity consumption has grown by 20-40\% annually in recent years, reaching 1-1.3\% of global electricity demand and contributing 1\% of energy-related greenhouse gas emissions in 2022~\cite{hintemann2022cloud}. 
However the contribution of the AI sector specifically towards these figures is unclear.

Recent work documenting the environmental impacts of ML has focused largely on quantifying the operational energy and carbon required to perform the training phase of the ML model life cycle~\cite{strubell2019energy, patterson2021carbon, dodge2022measuring, luccioni2023counting} due to the relative ease of measuring per-model energy use for that phase and the impressive quantity of energy required to perform a single training run~\cite{strubell2019energy,patterson2021carbon}. Yet, other phases of the ML model life cycle, such as inference, stand to impact the environment just as much, or more, than training due to the computational resources required to deploy modern models at scale. While inference on a single example requires much less computation than that required to train the same model, inference happens far more frequently than model training --- as many as billions of times a day for a model powering a popular user-facing product such as Google Translate.\footnote{Google reported translating more than 100 billion words per day in 2016, assuming an average query length of 100 words yields an estimate of 1 billion queries to the model per day. Source: \protect{\url{https://blog.google/products/translate/ten-years-of-google-translate/}}} Yet, in-depth work quantifying the costs of model inference and deployment is limited and their environmental impacts, in terms of energy and carbon as well as water and mining of rare earth minerals, have yet to be estimated. According to AWS, the largest global cloud provider, inference is estimated to make up 80 to 90\% of total ML cloud computing demand~\cite{barr2019amazon,nvidia-inference}, whereas a 2021 publication by Meta attributed approximately one-third of their internal end-to-end ML carbon footprint to model inference, with the remainder produced by data management, storage, and training~\cite{wu2021sustainable}; similarly, a 2022 study from Google attributed 60\% of its ML energy use to inference, compared to 40\% for training~\cite{patterson2022}. Given the increasing ubiquity of AI model deployment, it is crucial to go beyond these high-level statistics to get a better idea of the energy requirements and carbon emissions of model inference for different models and tasks.
In particular, looking at inference rather than training leads to drastically different conclusions when considering the multi-purpose (or ``general-purpose'') aspect specifically. Training a single model for multiple tasks can indeed be more energy-efficient when considering training costs only, but these gains can easily be lost and even reversed over the course of the model's lifetime, given how much inference is carried out when these models are deployed in user-facing applications like chat and web search.

To help shed light on this issue, we perform an extensive study measuring the amount of energy required to deploy various ML models and architectures, including large language models (LLMs)- as such, our study is, to our knowledge, the first to focus solely on the \emph{inference} phase of the ML model life cycle. We study 88 models across 10 tasks and 30 datasets, spanning applications in natural language and computer vision, analyzing the impact of end task, modality, model size, architecture, and learning paradigm (i.e. task-specific or multi-task/multi-purpose) on energy efficiency. We identify orders-of-magnitude differences in the amount of energy required per inference across models, modalities and tasks and shine light on an important trade-off between the benefit of multi-purpose systems, their energy cost, and ensuing carbon emissions.
By painting a more detailed picture of widely varying energy requirements for ML model inference, we hope this study can be useful for practitioners to better understand accuracy-efficiency trade-offs across tasks and models, as well as enabling better estimates, and projections and policy decisions at the sector level.


\section{Previous Work} \label{sec:previous-work}
Estimating the energy and emissions of ML models has remains a relatively under-explored topic, albeit one that has been gathering traction since Strubell et al's seminal article quantifying the energy and carbon emissions of a variety of then-large NLP models~\citeyearpar{strubell2019energy}. Since then, most studies have focused on estimating the energy consumed and carbon emitted during the training phase of neural networks -- this includes studies by Patterson et al.~\citeyearpar{patterson2021carbon,patterson2022}, who compared different models and analyzed factors influencing their emissions. There have also been studies of specific model architectures, e.g. BLOOM~\cite{luccioni2022estimating} and Nour~\cite{lakim2022holistic}, which carried out in-depth analyses of the different steps in the models' life cycle and their relative contribution towards the final quantity of carbon emissions.
Given the increasing deployment of ML models in the cloud, several studies have therefore looked at cloud-specific ways to reduce the emissions of ML models such as delayed scheduling, workload elasticity and choosing the least carbon-intensive electricity available~\citet{dodge2022measuring, hanafy2023carbonscaler,chien2023reducing}.

Despite these empirical studies, there is currently a lack of standardized methodology for quantifying and comparing the energy consumption and carbon emissions of ML models. There are several tools that exist, such as Code Carbon~\cite{schmidt2021codecarbon}, MLCO2~\cite{lacoste2019quantifying} and LLMCarbon~\cite{faiz2023llmcarbon}, all of which adopt different approaches and output different results (see~\cite{bannour2021evaluating} for a detailed comparison). It is therefore difficult to systematically compare the carbon footprints of different models. Existing tools and studies have also largely focused on the dynamic power consumption (i.e. the electricity necessary for powering hardware) and its resulting emissions. However, there have been several proposals to also take into account the embodied emissions of ML models (i.e. the emissions that can be attributed to the manufacturing of computing equipment) into carbon emissions estimates. This has been impeded by a lack of transparency from the designers of common computing hardware such as GPUs, although recent estimates have revealed that the embodied carbon footprint of an LLM trained and deployed on Meta's compute cluster constitutes up to 50\% of its carbon footprint~\cite{wu2021sustainable}.
While the majority of existing work has been focused on ML model training given that it is a more tractable part of the model life cycle (i.e. it is most often carried out over a set period of time on a specific compute instance), model inference has started to also become the subject of scholarship~\cite{desislavov2021compute,chien2023reducing}. Luccioni et al.'s study of BLOOM was the first of its kind to look at the specific energy costs related to deploying an LLM~\cite{luccioni2022estimating} and found that, over time, this can represent a significant portion of a model's overall carbon footprint.

The current study further pursues this line of work, delving deeper into the inference stage of ML models, the energy it consumes and the carbon it emits. By testing a variety of architectures on different tasks and datasets, we aim to gain a better understanding of the degree of variance that can be observed and how seemingly small user choices can result in large differences in model's environmental impacts.

\section{Methodology} \label{sec:methodology}

As stated above, our study focuses on the inference (i.e. deployment) stage in the model life cycle, aiming to address the knowledge gaps that currently exist with regards to its energy consumption and ensuing emissions. We describe how we chose the tasks, datasets and models in the sections below, and present the results of our analysis in Section~\ref{sec:results}.

\subsection{Task and dataset selection} \label{subsec:tasks-datasets}

As the starting point of our study, we chose 10 ML tasks from 5 different modalities: \texttt{Text-to-category} (text classification, token classification, extractive question answering), \texttt{Text-to-text} (masked language modeling, text generation, summarization), \texttt{Image-to-category} (image classification, object detection), \texttt{Image-to-text} (image captioning) and \texttt{Text-to-image} (image generation). These tasks were chosen because they are common in both Natural Language Processing and Computer Vision, allowing us to explore multiple modalities, and include several multimodal tasks (i.e. image captioning and image generation), allowing us to explore the nexus between several modalities as well.
To test each of the tasks listed above, we chose three of the most downloaded datasets from the Hugging Face Hub. We present the tasks and their corresponding datasets in Table~\ref{table:tasks-datasets}.

\begin{table}[h!]
\begin{tabular}{llll}
\hline
\multicolumn{1}{l}{\textbf{Task}} & \textbf{Datasets} & \textbf{Task} & \textbf{Datasets} \\ \toprule
\textbf{\begin{tabular}[c]{@{}l@{}}image \\ classification\end{tabular}} & \begin{tabular}[c]{@{}l@{}}CIFAR 10~\cite{cifar}\\ CIFAR 100~\cite{cifar}\\ ImageNet 1K~\cite{ILSVRC15} \end{tabular} & \textbf{\begin{tabular}[c]{@{}l@{}}question \\ answering\end{tabular}} & \begin{tabular}[c]{@{}l@{}}SQuAD\cite{squad}\\ SQuAD v2~\cite{squad2}\\ SciQ~\cite{SciQ}\end{tabular} \\ \midrule
\textbf{\begin{tabular}[c]{@{}l@{}}image\\ captioning\end{tabular}} & \begin{tabular}[c]{@{}l@{}}Visual Genome~\cite{visualgenome}\\ RedCaps~\cite{redcaps}\\ COCO~\cite{coco} \end{tabular} & \textbf{summarization} & \begin{tabular}[c]{@{}l@{}}SAMSum~\cite{samsum}\\ CNN-Daily Mail~\cite{cnn}\\ XSum~\cite{xsum}\end{tabular} \\ \midrule
\textbf{\begin{tabular}[c]{@{}l@{}}image\\ generation\end{tabular}} & \begin{tabular}[c]{@{}l@{}}DiffusionDB~\cite{diffusiondb}\\ ImageReward~\cite{imagereward}\\ SD Prompts~\cite{sdprompts}\end{tabular} & \textbf{\begin{tabular}[c]{@{}l@{}}text \\ classification\end{tabular}} & \begin{tabular}[c]{@{}l@{}}IMDB~\cite{imdb}\\ Rotten Tomatoes~\cite{rotten_tomatoes}\\ SST 2~\cite{sst}\end{tabular} \\ \midrule
\textbf{\begin{tabular}[c]{@{}l@{}}masked\\ language \\ modeling\end{tabular}} & \begin{tabular}[c]{@{}l@{}}BookCorpus~\cite{bookcorpus}\\ C4~\cite{c4}\\ OSCAR~\cite{oscar}\end{tabular} & \textbf{\begin{tabular}[c]{@{}l@{}}text \\ generation\end{tabular}} & \begin{tabular}[c]{@{}l@{}}WikiText~\cite{wikitext}\\ BookCorpus~\cite{bookcorpus}\\ OSCAR~\cite{oscar}\end{tabular} \\ \midrule
\textbf{\begin{tabular}[c]{@{}l@{}}object \\ detection\end{tabular}} & \begin{tabular}[c]{@{}l@{}}Visual Genome~\cite{visualgenome}\\ CPPE-5~\cite{cppe5}\\ COCO~\cite{coco}\end{tabular} & \textbf{\begin{tabular}[c]{@{}l@{}}token \\ classification\end{tabular}} & \begin{tabular}[c]{@{}l@{}}ReCoRD~\cite{superglue}\\ WikiANN~\cite{wikiann}\\ CoNLL 2003~\cite{conll2003}\end{tabular} \\ \bottomrule
\end{tabular}
\caption{A list of the tasks and datasets used in our study.}
\label{table:tasks-datasets}
\end{table}
\vspace{-25pt}

\subsection{Models} \label{subsec:tasks}

To be representative of a broad diversity of deployment use cases, we sampled 88 models, some of which were trained or finetuned specifically for the tasks that we selected, whereas others were designed to be used as zero-shot or multi-task models, to allow comparisons both for different architectures on a given task and between tasks for the same architecture.

\paragraph{Task-specific Models.} For all of the tasks listed above, we selected the 8 most popular models from the HuggingFace Hub (by number of downloads)~\footnote{We were obliged to discard some models, e.g. if they were trained on another language or if the specific task they were fine-tuned for was not compatible with any of the datasets selected.} - we present the full list of model identifiers in Table~\ref{table:all-finetuned} in the Supplementary Materials. For each model, we ran 1,000 inferences for each of the 3 datasets from the task it was trained for (listed in Table~\ref{table:tasks-datasets}),  using the Transformers~\cite{wolf2019huggingface} library. We ran each set of inferences 10 times to ensure statistical significance of our measurements. We set up the inferences sequentially -- i.e., without batching -- in order to reflect the variability of model deployment \emph{in situ}, which can make it difficult to batch model inputs.

\paragraph{Multi-Purpose Models.} In addition to the task-specific models listed above, we also selected 8 multi-purpose models to analyze on different tasks -- models that were specifically trained to perform well in various different application settings. We chose 4 sequence-to-sequence models of different sizes from the Flan-T5 family~\cite{chung2022scaling} (\texttt{base}, \texttt{large}, \texttt{xl} and \texttt{xxl}) and 4 decoder-only models from the BLOOMz family~\cite{muennighoff2022crosslingual}: \texttt{BLOOMz-560M}, \texttt{BLOOMz-1B}, \texttt{BLOOMz-3B} and \texttt{BLOOMz-7B}. We tested these on a subset of the tasks to allow a comparison of multi-purpose generative models with individual task-specific systems in terms of their energy consumption and emissions: question answering, text classification and summarization. We selected these three tasks because we were able to find a set of models that were capable of carrying them out with a unified model architecture (which wasn't possible for all tasks, especially ones that involved multiple modalities.) We prompted these 8 models in a zero-shot setting that was constant across models, e.g. \emph{"Summarize the following text: [text]. Summary:"} on the same 1,000 samples as the fine-tuned models, also repeating each experiment ten times to measure the significance of results.

We ran all of our experiments on a node of 8 NVIDIA A100-SXM4-80GB GPUs hosted on Amazon Web Services, and used the Code Carbon package~\cite{schmidt2021codecarbon} to measure both the energy consumed and the carbon emitted during inference~\footnote{While all of our experiments were run on a single GPU, the idle power usage of the other GPUs is also reflected in the numbers that we report in our results.}.
Given that all of our experiments were run in the same compute region (AWS's \texttt{us-west-2}), which is based in Oregon and has an average carbon intensity of 297.6 grams of $CO_2eq$ per kWh\footnote{The carbon intensity of an energy grid is measured in \emph{$CO_2eq$}, and not in $CO_2$ specifically, because the different greenhouse gases that are generated during electricity generation are reduced to a common denominator, that of carbon dioxide, or $CO_2$. For a more in-depth discussion of how this is done, see Luccioni and Hernandez-Garcia~\citeyearpar{luccioni2023counting}.}, this means that both the energy consumed during inference and the carbon emitted are correlated; we will therefore plot one or the other depending on which aspect of our results we are discussing. While the energy consumed during inference will remain similar for models deployed on A100 GPUs in other compute regions, the carbon emissions will vary depending on the source of energy used in the region -- it is therefore helpful to report both energy and carbon separately to allow for meaningful comparisons across regions and hardware.
We provide all the code used for our experiments in our \href{https://github.com/sashavor/co2_inference/}{GitHub repository}, alongside the logs produced by Code Carbon, which not only provides the total energy consumed but also a more fine-grained breakdown by hardware component (GPU, CPU and RAM), which can be used to carry out further analyses. In total, for all of model experimentation and evaluation, we used a total of 754.66 kWh of energy and emitted 178.97 kg of $CO_2eq$.

\section{Results} \label{sec:results}

We present our results in the subsections below: in Section~\ref{subsec:tasks-energy}, we analyze the range of energy used and carbon emitted for each task for task-specific models. In Section~\ref{subsec:zeroshot}, we shift our focus to multi-purpose (i.e. `zero-shot` models), looking at the variation between different sizes and architectures of multi-purpose models and the difference in the energy consumption and emissions between task-specific and multi-purpose models. In Section~\ref{subsubsection:training-inference}, we carry out a comparison between model training and inference costs for models of different sizes, calculating when parity is reached.

\subsection{Task-specific model analysis} \label{subsec:tasks-energy}


We start by analyzing the degree of variability in terms of the energy cost of ML models specifically trained for a variety of tasks.  Table~\ref{table:tasks-energy} shows each of the ten tasks that we analyzed as well as the mean energy used across all models for 1,000 inferences and its standard deviation. We can see that classification tasks for both images and text are on the lower end of the spectrum in terms of emissions (ranging between 0.002 and 0.007 kWh for 1,000 inferences), whereas generative tasks such as text generation and summarization use, on average, over 10 times more energy for the same number of inferences (around 0.05 kWh for 1,000 inferences), and multimodal tasks such as image captioning and image generation are on the highest end of the spectrum (0.06-2.9 kWh for 1,000 inferences). Text-based tasks are, all things considered, more energy-efficient than image-based tasks, with image classification requiring less energy (median of 0.0068 kWh for 1,000 inferences) than image generation (1.35 kWh) and, conversely, text generation (0.042 KwH) requiring more than text classification (0.0023 kWh). For comparison, charging the average smartphone requires  0.022 kWh of energy~\cite{EPA2024}, which means that the most efficient text generation model uses as much energy as 9\% of a full smartphone charge for 1,000 inferences, whereas the least efficient image generation model uses as much energy as 522 smartphone charges (11.49 kWh), or around half a charge per image generation~\footnote{Before January 2024, the \href{https://web.archive.org/web/20230903042020/https://www.epa.gov/energy/greenhouse-gases-equivalencies-calculator-calculations-and-references}{EPA website} estimated a smartphone charge to consume 0.012 kWh of energy, which was the number used for comparisons in an earlier version of this study.}, although there is also a large variation between image generation models, depending on the size of image that they generate.

\begin{table}[h!]
\begin{tabular}{l cc}
 & \multicolumn{2}{l}{\textbf{inference energy (kWh)}} \\ \toprule
\textbf{task} & \multicolumn{1}{c}{\textbf{mean}} & \multicolumn{1}{c}{\textbf{std}} \\ \midrule
text classification & \multicolumn{1}{r}{0.002} & 0.001 \\
extractive QA & \multicolumn{1}{r}{0.003} & 0.001 \\
masked language modeling & \multicolumn{1}{r}{0.003} & 0.001 \\
token classification & \multicolumn{1}{r}{0.004} & 0.002 \\
image classification & \multicolumn{1}{r}{0.007} & 0.001 \\
object detection & \multicolumn{1}{r}{0.038} & 0.02 \\
text generation & \multicolumn{1}{r}{0.047} & 0.03 \\
summarization & \multicolumn{1}{r}{0.049} & 0.01 \\
image captioning & \multicolumn{1}{r}{0.063} & 0.02 \\
image generation & \multicolumn{1}{r}{2.907} &  3.31 \\ \bottomrule
\end{tabular}
\caption{Mean and standard deviation of energy per 1,000 queries for the ten tasks examined in our analysis.}
\label{table:tasks-energy}
\end{table}
We can also observe that there is a large variation in the amount of energy used, from the least energy-intensive task, text classification, with mean consumption of 0.002 KwH per 1,000 inferences, to the most energy-intensive one, image generation, whose mean consumption is 2.9kWh. This means that the different models examined in our study can vary by a factor of over 1450 in terms of the energy required to perform the same number of inferences.
Intuitively, this is coherent given the decision space that different types of models have - from a binary classification task such as sentiment analysis (which can only output, for instance, a 0 for negative sentiment and a 1 for positive) to an entire vocabulary for text generation and summarization models.  The length of text generated also impacts energy usage: on average, text generation uses 15 times more energy than masked language modeling, which makes sense given that the masked language modeling task only generates a single token, whereas in our setup the text generation task generates 10 new tokens for each input text, with the length of the input text rising as new tokens are generated, since each sequence of tokens gets fed back into the model to generate subsequent tokens. Finally, for image-based tasks, the level of abstraction is lower and the decision space is larger given that they generate raw pixels as opposed to tokens for text, making image-based tasks more energy intensive than text based ones, e.g. image classification uses over 3 times more energy than text classification (0.007 vs. 0.002 kWh) and image generation uses, on average, over 60 times more energy than text generation (0.047 vs. 2.9 kWh).

\begin{figure}[h!]
  \centering
  \includegraphics[width=\linewidth]{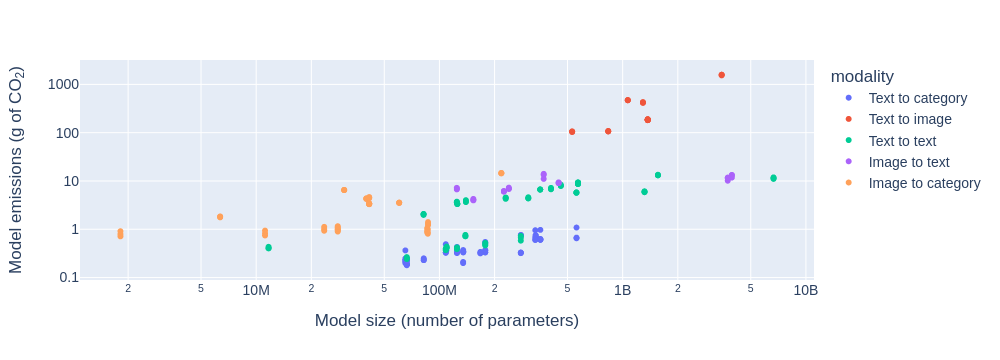}
  \caption{The 5 modalities examined in our study, with the number of parameters of each model on the x axis and the average amount of carbon emitted for 1000 inferences on the y axis. NB: Both axes are in logarithmic scale.}
  \label{fig:modality-energy}
\end{figure}

Next, we examine the respective influences of model size and task structure on model emissions. Figure~\ref{fig:modality-energy} shows the relationship between model emissions (in grams of $CO_2eq$ per 1,000 inferences) and sizes (in terms of the number of parameters) across the task categories listed in Section~\ref{subsec:tasks-datasets}.
We do observe a relationship between model size and quantity of emissions produced during inference, with differing progressions for each modality -- however, the task structure accounts for more of the variation than the model size does. We can observe once again that text-to-image is by far the most carbon- and energy-intensive task, with smaller image generation models such as \texttt{segmind/tiny-sd} that have around 500M parameters producing magnitudes more carbon than text-to-category models (100g vs. 0.6g of $CO_2eq$ per 1,000 inferences). Within the text-to-text tasks, we see two separate sets of models: the masked language modeling task following a lower trend, producing emissions akin to text-to-category models, compared to text generation and summarization tasks, which produce similar amounts of carbon to the image captioning models with a similar number of parameters.  For context, the most carbon-intensive image generation model (\texttt{stable-diffusion-xl-base-1.0}) generates 1,594 grams of $CO_2eq$ for 1,000 inferences, which is roughly the equivalent to 4.1 miles driven by an average gasoline-powered passenger vehicle~\cite{EPA2024}, whereas the least carbon-intensive text generation model (\texttt{distilbert-base-uncased}) generates as much carbon as 0.0006 miles driven by a similar vehicle, i.e. 6,833 times less. This can add up quickly when image generation models such as \href{https://openai.com/dall-e-2}{Dall{\textperiodcentered}E} and \href{https://www.midjourney.com/explore}{MidJourney} are deployed in user-facing applications and used by millions of users globally (we discuss this point further in Section~\ref{sec:discussion}).

The (high-level) takeaway of this analysis is that even for models specifically trained to carry out a single task, there is a large level of variation both within each task and an even larger one between tasks from different modalities. In essence, tasks that map both image and text inputs to categorical outputs are less energy- and carbon-intensive than those that generate text or images. Making these distinctions can help inform policies seeking to mitigate the environmental impacts of AI, given that it is important to be aware of this variation, which can sometimes reach several orders of magnitude. In the next section, we delve deeper into multi-purpose systems, which are meant to carry out several tasks concurrently, to better understand their environmental impacts and how they compare to task-specific models.

\subsection{The environmental cost of multi-purpose systems} \label{subsec:zeroshot}

The second part of our analysis examines multi-task models of two types: decoder only, from the BLOOMz family, and sequence-to-sequence models from the FLAN-T5 family, with the goal of comparing energy intensity and carbon emissions of models with differing numbers of parameters when applied to different tasks. To address this question, we selected a subset of 3 tasks -- text classification, extractive question answering, and summarization -- given their diversity and broad applicability in a variety of settings, and compare the 8 zero-shot models of different sizes, based on the same 3 datasets per task as described in Table~\ref{table:tasks-datasets}.

\subsubsection{Emissions of task-specific and multi-task architectures} \label{subsec-finetuned-zeroshot} \hfill\\


To start our analysis, we examined how the choice of model and architecture type impacts emissions given a specific task and dataset. For this analysis, we took the same 8 task-specific models described in Section~\ref{subsec:tasks} and compared their emissions to the 8 multi-purpose models described above.

\begin{figure}[h!]
  \centering
  \includegraphics[width=\linewidth]{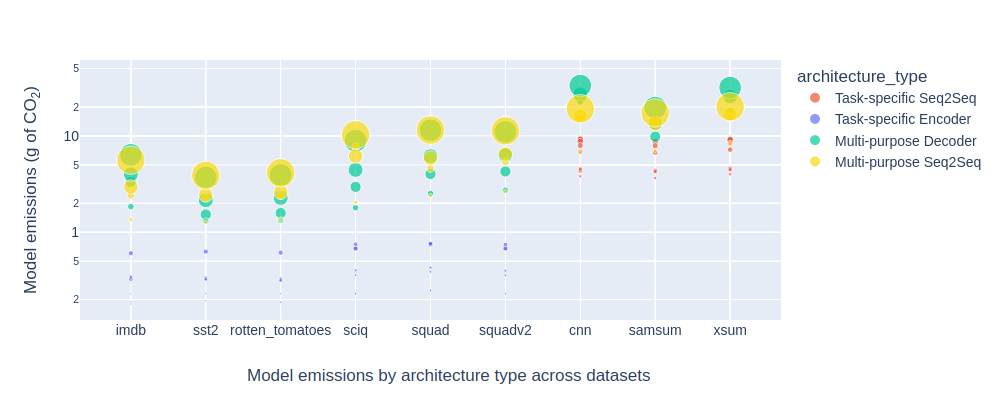}
  \caption{Model emissions (measured in g $CO_2eq$) and architecture type for each of the datasets from our analysis.   The y axis is in logarithmic scale, dot size is proportional to model size.}
  \label{fig:architectures-carbon}
\end{figure}

In Figure~\ref{fig:architectures-carbon}, we plot the mean query emissions for each model on a dataset-by-dataset basis. We can see that for the two \emph{discriminative} tasks, sentiment analysis (which includes SST 2, Rotten Tomatoes and IMDB datasets) and question answering (which encompasses SciQ, SQuAD and SQuAD v2) there is a clear distinction between task-specific discriminative models (in blue), which have less emissions than both multi-purpose sequence-to-sequence  (in yellow) and decoder-only generative models (in green). Given that the y axis in Figure~\ref{fig:architectures-carbon} is in logarithmic scale, this indicates that the difference is several orders of magnitude - e.g. with the most efficient task-specific models emit 0.3g of $CO_2eq$ per 1,000 inferences for extractive question answering on a dataset like SciQ, multi-purpose models emit 10g for the same task. This result follows intuitions derived from the model structures: while a task-specific model trained on binary text classification will carry out a softmax on a two-category vector to predict a class, a multi-purpose model will \emph{generate} `positive' or `negative', which logically requires more energy because the prediction is based on the model's entire vocabulary.
For the \emph{generative} task, summarization (represented by the SAMsum, XSum and CNN-Daily Mail datasets), the task-specific and multi-purpose models are closer in terms of emissions: task-specific sequence-to-sequence models generate 4-10g of $CO_2eq$ for 1,000 inferences, while multi-purpose models emit 20-30g for the same task. The difference appears to mostly come from model size -- all of the task-specific summarization models we looked at were 600 million parameters at most, compared to the larger multi-purpose architectures, which attained the 11 billion parameters.

We also carry out an evaluation of both the task-specific and multi-purpose models examined in our study to ensure that they have comparable performance. For task-specific models, we used the \texttt{evaluate} library~\cite{von2022evaluate} and the LM Evaluation Harness~\cite{eval-harness} for zero-shot models. Fundamentally speaking, it is hard to compare task-specific and multi-purpose models using the same metrics, given that task-specific models have a much more constrained decision space (e.g. two classes in the case of binary text classification), whereas multi-purpose models have a large output vocabulary to choose from, and are dependent upon the prompt schema and prompting strategy used. However, by utilizing two standardized packages (\texttt{evaluate} and \texttt{lm-evaluation-harness}) and keeping the prompting approach stable across zero-shot models, we endeavor to standardize our evaluation approach as much as possible.

\begin{figure}[h!]
  \centering
  \includegraphics[width=\linewidth]{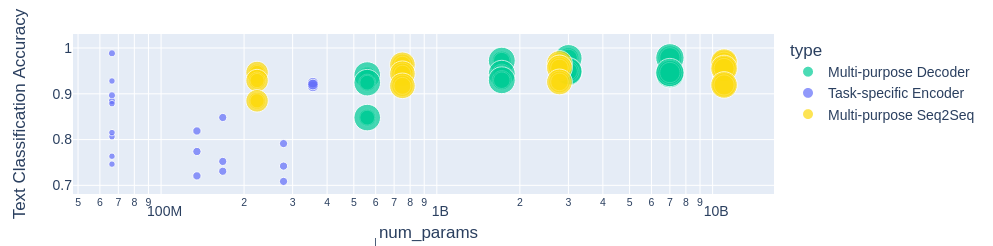}
  \caption{Model size, measured in number of parameters (x axis, logarithmic scale) and text classification accuracy (y axis), with dot size indicating the quantity of emissions (logarithmic scale).}
  \label{fig:architectures-accuracy}
\end{figure}

We hone in on one specific task, text classification, in Figure~\ref{fig:architectures-accuracy}, which illustrates the relationship between model size (x axis, in logarithmic scale), accuracy (y axis) and emissions (dot size, in logarithmic scale). Among task-specific encoder models, we observe that accuracy varies more widely, i.e. there are several smaller models of similar size and comparably small amounts of carbon emissions, with widely varying levels of accuracy. The multi-purpose models vary less in terms of accuracy, having higher average accuracy overall. Both sequence-to-sequence and decoder-only models produce comparable amounts of emissions (several orders of magnitude more than task-specific models).
We can see that mid-size multi-purpose models (in the 3B parameter range) may have slightly better accuracy compared to both larger and smaller models. However, given the many caveats and specificities involved in multi-purpose LLM evaluation, this difference may not be significant. We present the full results of our evaluation, which include the other 2 tasks, in Section~\ref{sec:model-eval} in the Supplementary Materials. 

\clearpage

\subsubsection{Differences within multi-purpose architectures} \label{subsec:zeroshot-diff} \hfill\\

\begin{figure}[h!]
  \centering
  \includegraphics[width=\linewidth]{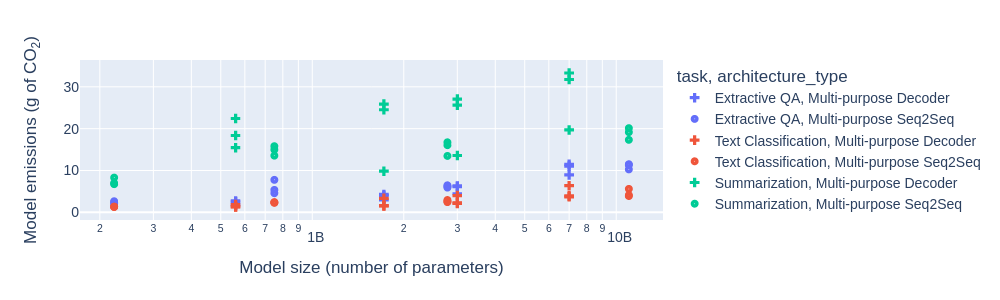}
  \caption{A plot of the total emissions (in grams of $CO_2eq$) for 1,000 inferences for all multi-purpose models.}
  \label{fig:zeroshot-carbon}
\end{figure}

Beyond the differences between task-specific and multi-purpose models generally, we also observed variation within the multi-purpose models that we examined. We present our results in Table~\ref{table:zeroshot}; in it, we can observe that on a per-architecture basis (i.e. within the family of decoder-only models and the family of sequence-to-sequence models), size and emissions are correlated, with smaller models emitting less carbon and using less energy. However, sequence-to-sequence models are more efficient than their decoder-only counterparts when models of the same size are compared: for instance, Flan-T5-XL and BLOOMz-3B are both of a similar size (around 3B parameters), but the former generates, on average, 2 grams of emissions less for 1,000 inferences than the latter. This difference holds when comparing Flan-T5-XXL, which is the biggest model in terms of parameter count in the multi-purpose models that we tested (11 billion), yet it has lower emissions (11.48g on average) compared to the smaller BLOOMz-7B. Comparing the models on a per-task basis in Figure~\ref{fig:zeroshot-carbon}, we can see the same pattern for zero-shot models as for task-specific ones, with text classification a less carbon-intensive task compared to question answering, and summarization the most intensive one of the three. The spread between the tasks is smaller for sequence-to-sequence models (indicated with dots in Figure~\ref{fig:zeroshot-carbon}), whereas for decoder-only models (indicated with crosses), the difference between the different tasks is more significant.

\begin{table}[h!]
\begin{tabular}{llll|llll}
\multicolumn{4}{c}{\textbf{seq2seq models}} & \multicolumn{4}{c}{\textbf{decoder-only models}} \\ \toprule
\multicolumn{1}{c}{\textbf{\begin{tabular}[c]{@{}c@{}}model\\ name\end{tabular}}} & \multicolumn{1}{c}{\textbf{\begin{tabular}[c]{@{}c@{}}number of \\ parameters \end{tabular}}} & \multicolumn{1}{c}{\textbf{\begin{tabular}[c]{@{}c@{}} emissions \\ (g $CO_2eq$) \end{tabular}}} &  \multicolumn{1}{c}{\textbf{\begin{tabular}[c]{@{}c@{}}energy \\ (kWh) \end{tabular}}} & \multicolumn{1}{c}{\textbf{\begin{tabular}[c]{@{}c@{}}model\\ name\end{tabular}}} & \multicolumn{1}{c}{\textbf{\begin{tabular}[c]{@{}c@{}} number of \\ parameters \end{tabular}}} & \multicolumn{1}{c}{\textbf{\begin{tabular}[c]{@{}c@{}} emissions \\ (g $CO_2eq$)\end{tabular}}} &  \multicolumn{1}{c}{\textbf{\begin{tabular}[c]{@{}c@{}}energy \\ (kWh) \end{tabular}}} \\ \midrule
\multicolumn{1}{l}{\textbf{Flan-T5-base}} & \multicolumn{1}{c}{222M} & \multicolumn{1}{c}{3.67}  &  \multicolumn{1}{c}{0.026} & \multicolumn{1}{l}{\textbf{BLOOMz-560M}} & \multicolumn{1}{c}{559M} & \multicolumn{1}{c}{7.5}  & \multicolumn{1}{c}{0.054} \\
\multicolumn{1}{l}{\textbf{Flan-T5-large}} & \multicolumn{1}{c}{750M} & \multicolumn{1}{c}{7.68} &\multicolumn{1}{c}{0.055} & \multicolumn{1}{l}{\textbf{BLOOMz-1B}} & \multicolumn{1}{c}{1.7B} &\multicolumn{1}{c}{8.66}  & \multicolumn{1}{c}{0.062} \\
\multicolumn{1}{l}{\textbf{Flan-T5-xl}} & \multicolumn{1}{c}{2.8B} & \multicolumn{1}{c}{8.08}  & \multicolumn{1}{c}{0.058} & \multicolumn{1}{l}{\textbf{BLOOMz-3B}} & \multicolumn{1}{c}{3B} & \multicolumn{1}{c}{10.17} & \multicolumn{1}{c}{0.073} \\
\multicolumn{1}{l}{\textbf{Flan-T5-xxl}} & \multicolumn{1}{c}{11B} & \multicolumn{1}{c}{11.48} & \multicolumn{1}{c}{0.083} & \multicolumn{1}{l}{\textbf{BLOOMz-7B}} & \multicolumn{1}{c}{7B} & \multicolumn{1}{c}{14.46}  & \multicolumn{1}{c}{0.104} \\ \bottomrule
\end{tabular}
\caption{Zero-shot models in our analysis with their architecture type, model size (in number of parameters), average quantity of emissions (in g of $CO_2eq$) and average energy usage (in kWh) for 1,000 inferences.}
\label{table:zeroshot}
\end{table}

We can analyse the relationship between sequence-to-sequence and decoder-only models noted in Table~\ref{table:zeroshot}: whereas for tasks such as summarization, decoder models do generate more emissions than sequence-to-sequence models of a similar size, for question answering and text classification, the two architectures have similar emissions.
This can again be explained by the differences in the model structures, specifically the attention mechanism: while sequence-to-sequence models only attend to the last layer of the input when producing their answers, decoder-only architectures attend to all layers for the full sequence -- leading to a stronger dependency on the output length for the number of operations, resulting in more emissions for tasks with longer outputs.

\begin{table}[]
\begin{tabular}{c|r|r|r|r|r|r|r|r|r}
\multicolumn{1}{l|}{} & \multicolumn{1}{l|}{} & \multicolumn{1}{l|}{\textbf{\begin{tabular}[c]{@{}l@{}}BLOOMz\\ 560M\end{tabular}}} & \multicolumn{1}{l|}{\textbf{\begin{tabular}[c]{@{}l@{}}BLOOMz\\ 1B\end{tabular}}} & \multicolumn{1}{l|}{\textbf{\begin{tabular}[c]{@{}l@{}}BLOOMz\\ 3B\end{tabular}}} & \multicolumn{1}{l|}{\textbf{\begin{tabular}[c]{@{}l@{}}BLOOMz\\ 7B\end{tabular}}} & \multicolumn{1}{l|}{\textbf{\begin{tabular}[c]{@{}l@{}}Flan-T5\\ base\end{tabular}}} & \multicolumn{1}{l|}{\textbf{\begin{tabular}[c]{@{}l@{}}Flan-T5\\ large\end{tabular}}} & \multicolumn{1}{l|}{\textbf{\begin{tabular}[c]{@{}l@{}}Flan-T5\\ xl\end{tabular}}} & \multicolumn{1}{l}{\textbf{\begin{tabular}[c]{@{}l@{}}Flan-T5\\ xxl\end{tabular}}} \\ \hline
\textbf{dataset} & \multicolumn{1}{c|}{\textbf{input}} & \multicolumn{1}{c|}{\textbf{output}} & \multicolumn{1}{c|}{\textbf{output}} & \multicolumn{1}{c|}{\textbf{output}} & \multicolumn{1}{c|}{\textbf{output}} & \multicolumn{1}{c|}{\textbf{output}} & \multicolumn{1}{c|}{\textbf{output}} & \multicolumn{1}{c|}{\textbf{output}} & \multicolumn{1}{c}{\textbf{output}} \\ \hline
\textbf{IMDB} & 58.73 & \cellcolor[HTML]{F9F9F9}1.64 & \cellcolor[HTML]{EAEAEA}2.61 & \cellcolor[HTML]{E2E2E2}1.72 & \cellcolor[HTML]{CCCCCC}1.53 & \cellcolor[HTML]{FFFFFF}1.00 & \cellcolor[HTML]{F3F3F3}1.00 & \cellcolor[HTML]{EEEEEE}1.00 & \cellcolor[HTML]{D1D1D1}1.00 \\ \hline
\textbf{\begin{tabular}[c]{@{}c@{}}Rotten \\ Tomatoes\end{tabular}} & 30.08 & \cellcolor[HTML]{FFFFFF}1.00 & \cellcolor[HTML]{FCFCFC}0.99 & \cellcolor[HTML]{F5F5F5}1.03 & \cellcolor[HTML]{E3E3E3}1.00 & \cellcolor[HTML]{FFFFFF}1.00 & \cellcolor[HTML]{F3F3F3}1.00 & \cellcolor[HTML]{F1F1F1}1.00 & \cellcolor[HTML]{E1E1E1}1.00 \\ \hline
\textbf{SST 2} & 28.35 & \cellcolor[HTML]{FFFFFF}0.98 & \cellcolor[HTML]{FDFDFD}0.99 & \cellcolor[HTML]{F6F6F6}1.01 & \cellcolor[HTML]{E5E5E5}1.02 & \cellcolor[HTML]{FFFFFF}1.00 & \cellcolor[HTML]{F5F5F5}1.00 & \cellcolor[HTML]{F3F3F3}1.00 & \cellcolor[HTML]{E3E3E3}1.00 \\ \hline
\textbf{SciQ} & 113.12 & \cellcolor[HTML]{FAFAFA}1.28 & \cellcolor[HTML]{EDEDED}1.25 & \cellcolor[HTML]{DEDEDE}1.10 & \cellcolor[HTML]{C7C7C7}1.10 & \cellcolor[HTML]{F7F7F7}2.03 & \cellcolor[HTML]{C9C9C9}5.41 & \cellcolor[HTML]{CCCCCC}3.12 & \cellcolor[HTML]{C5C5C5}2.42 \\ \hline
\textbf{SQuAD} & 134.00 & \cellcolor[HTML]{F2F2F2}1.93 & \cellcolor[HTML]{E2E2E2}1.96 & \cellcolor[HTML]{CCCCCC}2.02 & \cellcolor[HTML]{C2C2C2}1.95 & \cellcolor[HTML]{F4F4F4}2.01 & \cellcolor[HTML]{DCDCDC}2.15 & \cellcolor[HTML]{CECECE}2.16 & \cellcolor[HTML]{C2C2C2}2.13 \\ \hline
\textbf{SQuAD 2} & 115.85 & \cellcolor[HTML]{F0F0F0}2.33 & \cellcolor[HTML]{DFDFDF}2.54 & \cellcolor[HTML]{CCCCCC}2.58 & \cellcolor[HTML]{C3C3C3}2.41 & \cellcolor[HTML]{F1F1F1}2.28 & \cellcolor[HTML]{D4D4D4}2.74 & \cellcolor[HTML]{CCCCCC}2.71 & \cellcolor[HTML]{C3C3C3}2.58 \\ \hline
\textbf{CNN} & 54.00 & \cellcolor[HTML]{AEAEAE}12.05 & \cellcolor[HTML]{AAAAAA}11.91 & \cellcolor[HTML]{A5A5A5}11.73 & \cellcolor[HTML]{999999}10.34 & \cellcolor[HTML]{CBCBCB}8.52 & \cellcolor[HTML]{BCBCBC}11.34 & \cellcolor[HTML]{BABABA}11.34 & \cellcolor[HTML]{B4B4B4}10.68 \\ \hline
\textbf{SamSUM} & 47.82 & \cellcolor[HTML]{BBBBBB}9.54 & \cellcolor[HTML]{C5C5C5}9.41 & \cellcolor[HTML]{BEBEBE}9.75 & \cellcolor[HTML]{B3B3B3}9.85 & \cellcolor[HTML]{CBCBCB}10.56 & \cellcolor[HTML]{BEBEBE}11.05 & \cellcolor[HTML]{BFBFBF}10.18 & \cellcolor[HTML]{B7B7B7}10.57 \\ \hline
\textbf{XSum} & 53.85 & \cellcolor[HTML]{B5B5B5}11.53 & \cellcolor[HTML]{A7A7A7}12.22 & \cellcolor[HTML]{A8A8A8}11.94 & \cellcolor[HTML]{9C9C9C}11.92 & \cellcolor[HTML]{C8C8C8}12.95 & \cellcolor[HTML]{BABABA}13.62 & \cellcolor[HTML]{B8B8B8}13.49 & \cellcolor[HTML]{B2B2B2}13.09
\end{tabular}
\caption{Average input and output length (in number of tokens) for the 8 zero-shot models and 9 tasks examined as part of our study. The darker the cell, the more carbon was output by the model for the task.}
\label{table:input_output}
\end{table}

\begin{figure}[h!]
  \centering
  \includegraphics[width=\linewidth]{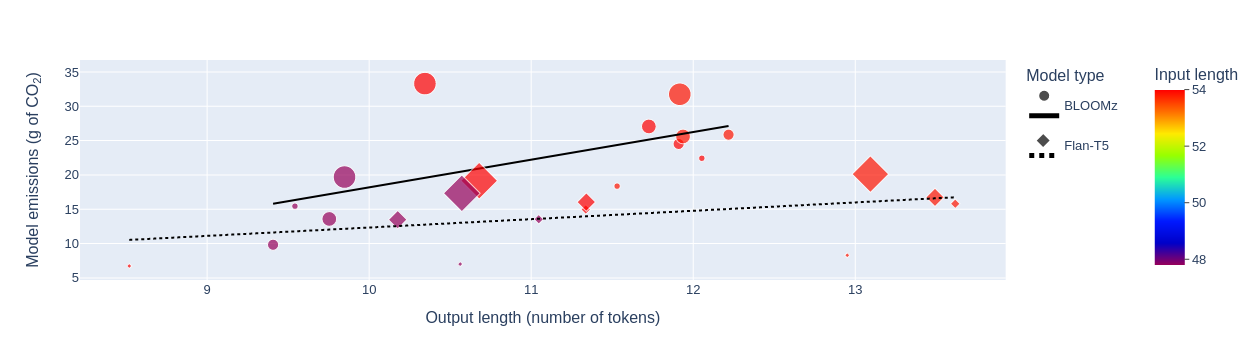}
  \caption{A plot of the output length (X axis) and carbon emissions (Y axis) for the summarization task. The symbol refers to the type of architecture (BLOOMz vs Flan-T5), symbol size references the relative model size (in terms of the number of parameters), and color the input length.}
  \label{fig:summarization-outputs}
\end{figure}

We further verify this intuition in  Table~\ref{table:input_output} and Figure~\ref{fig:summarization-outputs}: while there is some variation between models and datasets in Table~\ref{table:input_output}, the distribution of output lengths is consistent with our expectations for the different task categories: tasks with longer outputs result in more emissions, especially for decoder-only models. Figure~\ref{fig:summarization-outputs} delves further into the relationship between average output length, carbon emissions, and model structures for the different summarization datasets. It shows a clear correlation between output length and measured emissions, with a higher slope for the decoder-only architectures (the BLOOMz family of models) than for the sequence-to-sequence architectures (the Flan-T5 family).


As we have observed in the current section, there is no `one-size-fits-all' pattern for multi-purpose models either -- they too exhibit variation in terms of their emissions and energy usage, which can be attributed to different factors, including model size and output length. This would indicate that more careful consideration is needed when making choices to deploy these models for different tasks and applying them in different scenarios. We further discuss our results and further avenues of research in the next and final section.

\subsection{Comparing model training and inference costs} \label{subsubsection:training-inference}

An important trade-off for many AI practitioners and policy-makers is determining when exactly model inference costs reach parity with model training (and fine-tuning) - i.e. when does the \emph{deployment} of models use as much energy as their initial \emph{training}? This comparison is often hard to make because it requires the total energy cost of all steps of the ML model life cycle, which is very rarely available. Of the models that we examined in our study, neither the BLOOMz nor the Flan-T5 families of models reported the total energy used nor carbon emitted during their training in the papers describing the models. However, given that the BLOOMz models are fine-tuned versions of the original BLOOM family of models~\cite{workshop2022bloom}, we can base ourselves on the \href{https://github.com/bigscience-workshop/carbon-footprint/}{logs} provided by the authors of the BLOOM carbon footprint estimation paper~\cite{luccioni2022estimating}. We can add to these numbers the energy cost of fine-tuning each model, which we were able to estimate based on the training logs provided by the authors of the BLOOMz paper~\cite{muennighoff2022crosslingual}, although we were lacking the necessary information to infer the carbon footprint~\footnote{The energy consumption can be based on the Thermal Design Power (TDP) of the GPUs used -- while it assumes 100\% GPU utilization, it is the most accurate estimate possible without energy usage tracking during training.}. We present these numbers, alongside the average energy consumption per inference, in Table~\ref{table:training-inference}. We can see that the amount of energy required per inference varies from 5.4 × $10^{-5}$ for the smallest model, BLOOMz-560M to 1.0 × $10^{-4}$ kWh for the biggest one, BLOOMz-7B. This is coherent to the numbers reported by Luccioni et al. for BLOOM-176B, which required, on average, 0.004 kWh of energy per query, or 40 times more than BLOOMz-7B, being roughly 25 times bigger~\cite{luccioni2022estimating} - although this included API deployment of the model, which is not the case for the models in our study.

\begin{table}[h!]
\begin{tabular}{lrrrr}
 & \multicolumn{1}{l}{\textbf{BLOOMz-7B}} & \multicolumn{1}{l}{\textbf{BLOOMz-3B}} & \multicolumn{1}{l}{\textbf{BLOOMz-1B}} & \multicolumn{1}{l}{\textbf{BLOOMz-560M}} \\ \toprule
\textbf{Training energy (kWh)} & 51,686 & 25,634 & 17,052 & 10,505 \\
\textbf{Finetuning energy (kWh)} & 7,571 & 3,242 & 1,081 & 543 \\
\textbf{Inference energy (kWh)} & 1.0 × $10^{-4}$ &	7.3 × $10^{-5}$ & 6.2 × $10^{-5}$ & 5.4 × $10^{-5}$ \\
\textbf{\begin{tabular}[c]{@{}l@{}} Cost parity (\# inferences) \end{tabular}} & 592,570,000 & 395,602,740 & 292,467,741 & 204,592,592 \\ \bottomrule
\end{tabular}
\caption{The BLOOMz models from our study with their training energy cost (from~\cite{luccioni2022estimating}), finetuning energy cost (from~\cite{muennighoff2022crosslingual}), inference cost (from the present study), and cost parity, as the number of inferences required to sum to the training cost.}
\label{table:training-inference}
\end{table}

If we compare the amount of energy used per inference for each of the models with the total amount of energy used for both training and fine-tuning them, we can estimate how many inferences would be needed to be carried out with a given model in order for the cost of inference to reach the cost of training. As can be seen in Table~\ref{table:training-inference}, this varies depending on model size: from around 200 million inferences for the smallest model, BLOOMz-560M, to over 590 million inferences for the biggest model, BLOOMz-7B. This may seem like a lot if a single instance of a model is deployed, but can add up quickly if there are multiple instances of models deployed in parallel. For instance, it has been estimated that, at its peak, ChatGPT had upward of 10 million users per day~\cite{oremus2023ai}; the most recent statistics indicate that the ChatGPT login page received 1.7B visits in October 2023~\footnote{According to SimilarWeb: \url{https://www.similarweb.com/website/chat.openai.com/}.}. Even assuming a single query per user, which is rarely the case, the energy costs of deploying it would surpass its training costs after a few weeks or months of deployment.

While the BLOOMz models are not deployed in real-time in the same manner as ChatGPT, they have been downloaded hundreds of thousands of times from the Hugging Face Hub, which would indicate that they have been extensively used by the open-source community: at the time of writing this article (November 2023), BLOOMz-7B has been downloaded 606,096 times, BLOOMz-3B has been downloaded 357,368 times, BLOOMz-1B has been downloaded 61,757 times and BLOOMz-560m has been downloaded 498,601 times. They have also been finetuned for a number of downstream tasks, such as chat, and deployed in HuggingFace Spaces, interactive interfaces for model interaction. While this analysis represents a relatively small sample of models,
analyses such as this are vital for estimating the relative energy consumption (and ensuing emissions) of different stages of the ML training and deployment cycle, understanding trade-offs between training and inference emissions patterns, and characterizing the lifetime emissions of ML models, and we hope that others will be possible in the future, which would require more transparency from model creators regarding both the up front (i.e. training) and downstream (i.e. inference) costs of ML models. We discuss the importance of transparency and other important actions that members of the community can take in the next, and final, section.

\section{Discussion} \label{sec:discussion}
There have been limited studies regarding the energy consumption and carbon emissions of LLM inference, largely due to its distributed nature --- compared to the relatively time- and location-constrained nature of training --- making it difficult to make meaningful comparisons between different models and tasks. In this work, we have endeavored to keep as many parameters stable as possible, including the code, hardware, datasets, batch size and Python library. We provide all of the \href{https://github.com/sashavor/co2_inference/}{code} that we used for our analysis as well as an \href{https://huggingface.co/spaces/sasha/CO2_inference}{interactive tool} to allow users to more deeply explore the results we present here. We also highlight the main high-level takeaways of our study below:

\paragraph{Generative tasks are more energy- and carbon-intensive compared to discriminative tasks} As shown in Figure~\ref{fig:tasks-carbon}, the most energy- and carbon-intensive tasks are those that generate new content: text generation, summarization, image captioning, and image generation.

\paragraph{Tasks involving images are more energy- and carbon-intensive compared to those involving text alone} More specifically, tasks involving predicting categories (text-to-category, image-to-category) are less energy-intensive than those involving generating images (e.g. text-to-image), with those involving text between the two (see Figure~\ref{fig:modality-energy}).

\paragraph{Decoder-only models are slightly more energy- and carbon- intensive than sequence-to-sequence models for models of a similar size and applied to the same tasks} The findings we present in Table~\ref{table:zeroshot},  Figure~\ref{fig:architectures-carbon}, and Figure~\ref{fig:summarization-outputs} would indicate that more computation (i.e. energy) is required for decoder-only tasks, and that this phenomenon is particularly marked for tasks with longer outputs.
This observation is worth verifying for other architectures from both categories, and well as other tasks and datasets.

\paragraph{Training remains orders of magnitude more energy- and carbon- intensive than inference} We have provided initial numbers for comparing the relative energy costs of model training, finetuning and inference for different sizes of models from the BLOOMz family, and found that the parity between training/finetuning and inference grows with model size. While the ratio is hundreds of millions of inferences for a single training, given the ubiquity of ML model deployment, this parity can be reached quickly for many popular models.

\paragraph{Using multi-purpose models for discriminative tasks is more energy-intensive compared to task-specific models for these same tasks} This is especially the case for text classification (on IMDB, SST 2 and Rotten Tomatoes) and question answering (on SciQ, SQuAD v1 and v2), where the gap between task-specific and zero-shot models is particularly large, and less so for summarization (for CNN-Daily Mail, SamSUM and XSum). As can be seen in Table~\ref{table:input_output}, the difference between multi-purpose models and task-specific models is amplified as the length of output gets longer. \\

We find this last point to be the most compelling takeaway of our study, given the current paradigm shift away from smaller models finetuned for a specific task towards models that are meant to carry out a multitude of tasks at once, deployed to respond to a barrage of user queries in real time. This transition has been happening both in ML research since the advent of GPT-3~\cite{brown2020language}, which illustrated the potential for few- and zero-shot learning with language models, as well as in consumer settings, with LLMs such as GPT-4 and PaLM being deployed in user-facing products such as web search~\cite{Bing2023,Bard2023}, email, and navigation~\cite{Bard2023_2}, where smaller, task-specific versions of models such as BERT were previously used~\cite{Bing2019, Bert2019}. While it is hard to quantify the environmental impacts of this transition given the lack of transparency of technology companies regarding both the number of parameters, architecture and carbon emissions of their products, we can make a comparison based on the experiments carried out in the present study. For instance, the average emissions of a BERT-based model fine-tuned for extractive question answering (\texttt{bert-large-uncased-whole-word-masking-finetuned-squad}), a task akin to extractive web search, is 0.70g $CO_2eq$ per 1,000 queries, which is less than 3 times that of the multi-purpose models (2.36g for \texttt{Flan-T5 base} and 2.34g for \texttt{BLOOMz-560M}). The difference is much more drastic if comparing BERT-based models for tasks such as text classification with the larger multi-purpose models: for instance \texttt{bert-base-multilingual-uncased-sentiment} emits just 0.32g of $CO_2eq$ per 1,000 queries, compared to 2.66g for \texttt{Flan-T5-XL} and 4.67g for \texttt{BLOOMz-7B}. For comparison, the first PaLM model, released in 2022, has 540 billion parameters~\cite{chowdhery2022palm}, whereas GPT-3 has 175 billion parameters~\cite{brown2020language}~\footnote{The exact number of parameters of GPT-4 and PaLM 2 have not been publicly shared.}. While we see the benefit of deploying generative zero-shot models given their ability to carry out multiple tasks, we do not see convincing evidence for the necessity of their deployment in contexts where tasks are well-defined, for instance web search and navigation, given these models' energy requirements.


Finally, the intent of our study is to set the stage for better understanding of the energy requirements and carbon emissions of the final, often overlooked, step in the ML model life cycle: model deployment. The comparison between training, finetuning and inference energy requirements carried out in Section~\ref{subsubsection:training-inference} is, to our knowledge, the first comparison of its kind, and paves the way to a better understanding of how the different stages of an ML model's lifecycle add up in terms of energy use. These are important data points that can help inform both our fellow AI researchers and practitioners, as well as policy-makers who are working towards estimating and regulating the environmental impacts of AI models and ICT in general.  We recognize that our study is not representative of all deployment contexts and constraints -- our intent is to establish a set of initial data points and to set the stage for testing and comparing other models. In fact, our study highlights many potential avenues for future research aimed towards a better understanding of the myriad factors that influence the efficiency of inference, including the choice of architecture, the usage of techniques such as distillation, the number of parameters, the choice of hardware and the numerical (i.e. floating point) precision of model parameters. While we encourage continued work analysing open-source models, we note that the growing lack of transparency in model architecture and training details
makes this line of work, alongside many branches relating to fairness and accountability in machine learning, increasingly difficult to carry out. Given our findings and the increased deployment of generative, multi-purpose AI models, we hope that both ML researchers and practitioners will practice transparency regarding the nature and impacts of their models, to enable better understanding of their environmental impacts.
\clearpage
\section*{Ethical Considerations Statement}
The main ethical concerns that we faced in our experimentation is the sheer amount of energy needed and carbon emissions generated by our study, given that we ran each of the 88 models on 3 datasets 10 times to ensure statistical significance of our measurements. In total, for all of model experimentation and evaluation, we used a total of 754.66 kWh of energy and emitted 178.97 kg of $CO_2eq$. In order to reduce our impacts as much as possible, we did all up-front experimentations on smaller portions of the dataset (to reduce wasted resources).

\section*{Researcher Positionality Statement}
The authors of this paper have backgrounds in theoretical and applied machine learning and work in institutions based in North America. We therefore recognize that our way of planning and running experiments is not necessarily reflective of other institutions from other regions, or the constraints faced by researchers from institutions with more limited access to compute.

\section*{Adverse Impacts Statement}
We recognize that our work can be perceived as a critique of ML deployment in general, given the analysis that we provide of its environmental impacts. This could be used as an argument to stop pursuing ML research and development, or as a way of targeting specific companies or organizations. Our intention, however, is to shed additional light on the environmental impacts of ML, in order to help model developers and researchers make more informed choices as a function of their environmental footprint or energy usage.

\begin{acks}
We thank Will Alpine, Nima Boscarino, Priya Donti, Régis Pierrard, David Rolnick, Roy Schwartz and Rajiv Shah for their useful feedback and suggestions.
\end{acks}

\clearpage

\bibliographystyle{ACM-Reference-Format}
\bibliography{bibliography}

\clearpage
\appendix

\section{Full list of task-specific models tested}

\begin{table}[h!]
\small
\begin{tabular}{l|l|l|l}
\hline
\textbf{Task} & \textbf{Models} & \textbf{Task} & \multicolumn{1}{l|}{\textbf{Models}} \\ \hline
\textbf{\begin{tabular}[c]{@{}l@{}}image \\ classification\end{tabular}} & \begin{tabular}[c]{@{}l@{}}microsoft/resnet-50             \\ microsoft/beit-base-patch16-224\\ google/vit-base-patch16-384                  \\ facebook/convnextv2-tiny-22k-384           \\ microsoft/resnet-18                         \\ google/mobilenet\_v1\_0.75\_192               \\ facebook/convnextv2-tiny-1k-224            \\ google/vit-base-patch16-224\end{tabular} & \textbf{\begin{tabular}[c]{@{}l@{}}question \\ answering\end{tabular}} & \begin{tabular}[c]{@{}l@{}}distilbert-base-uncased-distilled-squad\\ distilbert-base-cased-distilled-squad\\ deepset/roberta-base-squad2\\ bert-large-uncased-whole-word-masking-finetuned-squad\\ timpal0l/mdeberta-v3-base-squad2\\ deepset/tinyroberta-squad2\\ deepset/electra-base-squad2\\ deepset/bert-large-uncased-whole-word-masking-squad2\end{tabular} \\ \hline
\textbf{\begin{tabular}[c]{@{}l@{}}image\\ captioning\end{tabular}} & \begin{tabular}[c]{@{}l@{}}nlpconnect/vit-gpt2-image-captioning \\ Salesforce/blip-image-captioning-large \\ Salesforce/blip-image-captioning-base\\ microsoft/git-large-coco          \\ Salesforce/blip2-flan-t5-xl      \\ Salesforce/blip2-opt-2.7b         \\ ydshieh/vit-gpt2-coco-en          \\ microsoft/git-base\end{tabular} & \textbf{summarization} & \begin{tabular}[c]{@{}l@{}}sshleifer/distilbart-xsum-12-6\\ sshleifer/distilbart-cnn-12-6\\ pszemraj/led-large-book-summary\\ google/pegasus-xsum\\ google/pegasus-large\\ google/pegasus-multi\_news\\ facebook/bart-large-cnn\\ ainize/bart-base-cnn\end{tabular} \\ \hline
\textbf{\begin{tabular}[c]{@{}l@{}}image\\ generation\end{tabular}} & \begin{tabular}[c]{@{}l@{}}runwayml/stable-diffusion-v1-5         \\ stabilityai/stable-diffusion-2-1        \\ stabilityai/stable-diffusion-xl-base-1.0   \\ CompVis/stable-diffusion-v1-4   \\ prompthero/openjourney  \\ dreamlike-art/dreamlike-photoreal-2.0 \\ nota-ai/bk-sdm-tiny   \\ segmind/tiny-sd \end{tabular} & \textbf{\begin{tabular}[c]{@{}l@{}}text \\ classification\end{tabular}} & \begin{tabular}[c]{@{}l@{}}distilbert-base-uncased-finetuned-sst-2-english\\ nlptown/bert-base-multilingual-uncased-sentiment\\ twitter-roberta-base-sentiment-latest\\ cardiffnlp/twitter-xlm-roberta-base-sentiment\\ lvwerra/distilbert-imdb\\ siebert/sentiment-roberta-large-english\\ finiteautomata/bertweet-base-sentiment-analysis\\ sbcBI/sentiment\_analysis\_mode\end{tabular} \\ \hline
\textbf{\begin{tabular}[c]{@{}l@{}}masked\\ language \\ modeling\end{tabular}} & \begin{tabular}[c]{@{}l@{}}bert-base-uncased  \\ xlm-roberta-base     \\ distilbert-base-uncased\\ roberta-base        \\ albert-base-v2            \\ bert-base-cased       \\ microsoft/deberta-base  \\ bert-base-multilingual-cased\end{tabular} & \textbf{\begin{tabular}[c]{@{}l@{}}text \\ generation\end{tabular}} & \begin{tabular}[c]{@{}l@{}}gpt2                      \\ bigscience/bloom-560m     \\ distilgpt2                \\ facebook/opt-6.7b         \\ EleutherAI/gpt-neo-125m  \\ gpt2-medium              \\ facebook/opt-1.3b      \\ gpt2-xl\end{tabular} \\ \hline
\textbf{\begin{tabular}[c]{@{}l@{}}object \\ detection\end{tabular}} & \begin{tabular}[c]{@{}l@{}}facebook/detr-resnet-50                    \\ hustvl/yolos-tiny                                   \\ jozhang97/deta-swin-large                        \\ facebook/detr-resnet-101              \\ hustvl/yolos-small                                \\ SenseTime/deformable-detr \\ polejowska/detr-r50-cd45rb-8ah-6l\\ polejowska/detr-r50-cd45rb-1ah-6l\end{tabular} & \textbf{\begin{tabular}[c]{@{}l@{}}token \\ classification\end{tabular}} & \begin{tabular}[c]{@{}l@{}}QCRI/bert-base-multilingual-cased-pos-english\\ dslim/bert-base-NER                 \\ dslim/bert-large-NER                  \\ Jean-Baptiste/roberta-large-ner-english  \\ oliverguhr/fullstop-punctuation-multilang-large   \\ Babelscape/wikineural-multilingual-ner      \\ ml6team/keyphrase-extraction-distilbert-inspec\\ obi/deid\_roberta\_i2b2\end{tabular} \\ \hline
\end{tabular}
\caption{The full list of the 80 finetuned models that were tested for the ten tasks we analyzed.}
\label{table:all-finetuned}
\end{table}



\clearpage

\section{Model Evaluation} \label{sec:model-eval}

\begin{table}[]
\begin{tabular}{l|l|l|l|l|l|l|l|l|l}
\textbf{model} & \textbf{\begin{tabular}[c]{@{}l@{}}SST 2 \\ (acc)\end{tabular}} & \textbf{\begin{tabular}[c]{@{}l@{}}IMDB \\ (acc)\end{tabular}} & \textbf{\begin{tabular}[c]{@{}l@{}}Rotten \\ Tomatoes \\ (acc)\end{tabular}} & \textbf{\begin{tabular}[c]{@{}l@{}}SciQ \\ (acc)\end{tabular}} & \textbf{\begin{tabular}[c]{@{}l@{}}SQuAD \\ (F1)\end{tabular}} & \textbf{\begin{tabular}[c]{@{}l@{}}SQuAD v2 \\ (F1, has \\ answer)\end{tabular}} & \textbf{\begin{tabular}[c]{@{}l@{}}SamSUM\\ (ROUGE)\end{tabular}} & \textbf{\begin{tabular}[c]{@{}l@{}}XSum\\ (ROUGE)\end{tabular}} & \textbf{\begin{tabular}[c]{@{}l@{}}CNN \\ (ROUGE)\end{tabular}} \\ \hline
\textbf{bloomz-560m} & 0.92 & 0.94 & 0.85 & 0.92 & 0.43 & 0.21 & 0.23 & 0.15 & 0.10 \\ \hline
\textbf{bloomz-1b7} & 0.94 & 0.97 & 0.93 & 0.96 & 0.50 & 0.25 & 0.26 & 0.16 & 0.18 \\ \hline
\textbf{bloomz-3b} & 0.95 & 0.98 & 0.95 & 0.97 & 0.53 & 0.26 & 0.28 & 0.17 & 0.21 \\ \hline
\textbf{bloomz-7b1} & 0.94 & 0.98 & 0.95 & 0.97 & 0.54 & 0.27 & 0.32 & 0.21 & 0.09 \\ \hline
\textbf{flan-t5-xxl} & 0.96 & 0.97 & 0.92 & 0.72 & 0.98 & 0.49 & 0.30 & 0.37 & 0.23 \\ \hline
\textbf{flan-t5-xl} & 0.96 & 0.97 & 0.93 & 0.66 & 0.97 & 0.49 & 0.49 & 0.38 & 0.24 \\ \hline
\textbf{flan-t5-large} & 0.94 & 0.96 & 0.92 & 0.53 & 0.97 & 0.50 & 0.45 & 0.30 & 0.24 \\ \hline
\textbf{flan-t5-base} & 0.93 & 0.95 & 0.88 & 0.61 & 0.95 & 0.48 & 0.46 & 0.32 & 0.23 \\ \hline
\textbf{\begin{tabular}[c]{@{}l@{}}distilbert-base-uncased\\ -distilled-squad\end{tabular}} &  &  &  & 0.44 & 0.87 & 0.86 &  &  &  \\ \hline
\textbf{\begin{tabular}[c]{@{}l@{}}distilbert-base-cased-\\ distilled-squad\end{tabular}} &  &  &  & 0.46 & 0.87 & 0.87 &  &  &  \\ \hline
\textbf{deepset/roberta-base-squad2} &  &  &  & 0.48 & 0.93 & 0.83 &  &  &  \\ \hline
\textbf{\begin{tabular}[c]{@{}l@{}}bert-large-uncased-whole-\\ word-masking-finetuned-squad\end{tabular}} &  &  &  & 0.48 & 0.93 & 0.84 &  &  &  \\ \hline
\textbf{\begin{tabular}[c]{@{}l@{}}timpal0l/mdeberta-v3-\\ base-squad2\end{tabular}} &  &  &  & 0.46 & 0.91 & 0.90 &  &  &  \\ \hline
\textbf{deepset/tinyroberta-squad2} &  &  &  & 0.45 & 0.98 & 0.91 &  &  &  \\ \hline
\textbf{deepset/electra-base-squad2} &  &  &  & 0.48 & 0.89 & 0.82 &  &  &  \\ \hline
\textbf{\begin{tabular}[c]{@{}l@{}}deepset/bert-large-uncased-\\ whole-word-masking-squad2\end{tabular}} &  &  &  & 0.46 & 0.92 & 0.92 &  &  &  \\ \hline
\textbf{sshleifer/distilbart-xsum-12-6} &  &  &  &  &  &  & 0.20 & 0.45 & 0.23 \\ \hline
\textbf{sshleifer/distilbart-cnn-12-6} &  &  &  &  &  &  & 0.29 & 0.21 & 0.44 \\ \hline
\textbf{\begin{tabular}[c]{@{}l@{}}pszemraj/led-large-\\ book-summary\end{tabular}} &  &  &  &  &  &  & 0.33 & 0.16 & 0.33 \\ \hline
\textbf{pegasus-xsum} &  &  &  &  &  &  & 0.22 & 0.22 & 0.22 \\ \hline
\textbf{pegasus-large} &  &  &  &  &  &  & 0.27 & 0.17 & 0.34 \\ \hline
\textbf{pegasus-multi\_news} &  &  &  &  &  &  & 0.12 & 0.16 & 0.29 \\ \hline
\textbf{facebook/bart-large-cnn} &  &  &  &  &  &  & 0.32 & 0.21 & 0.44 \\ \hline
\textbf{ainize/bart-base-cnn} &  &  &  &  &  &  & 0.27 & 0.16 & 0.26 \\ \hline
\textbf{\begin{tabular}[c]{@{}l@{}}distilbert-base-uncased-\\ finetuned-sst-2-english\end{tabular}} & 0.99 & 0.88 & 0.90 &  &  &  &  &  &  \\ \hline
\textbf{\begin{tabular}[c]{@{}l@{}}nlptown/bert-base-\\ multilingual-uncased-sentiment\end{tabular}} & 0.75 & 0.85 & 0.73 &  &  &  &  &  &  \\ \hline
\textbf{\begin{tabular}[c]{@{}l@{}}twitter-roberta-base-\\ sentiment-latest\end{tabular}} & 0.82 & 0.80 & 0.77 &  &  &  &  &  &  \\ \hline
\textbf{\begin{tabular}[c]{@{}l@{}}cardiffnlp/twitter-xlm-roberta-\\ base-sentiment\end{tabular}} & 0.79 & 0.71 & 0.74 &  &  &  &  &  &  \\ \hline
\textbf{lvwerra/distilbert-imdb} & 0.88 & 0.93 & 0.82 &  &  &  &  &  &  \\ \hline
\textbf{\begin{tabular}[c]{@{}l@{}}siebert/sentiment-roberta-\\ large-english\end{tabular}} & 0.92 & 0.92 & 0.92 &  &  &  &  &  &  \\ \hline
\textbf{\begin{tabular}[c]{@{}l@{}}finiteautomata/bertweet-\\ base-sentiment-analysis\end{tabular}} & 0.82 & 0.72 & 0.77 &  &  &  &  &  &  \\ \hline
\textbf{sbcBI/sentiment\_analysis\_model} & 0.81 & 0.75 & 0.76 &  &  &  &  &  &
\end{tabular}
\caption{Full performance metrics for the 32 models (24 finetuned, 8 multi-purpose) that we evaluated as part of our study.}
\label{table:evaluation}
\end{table}

\begin{figure}[h!]
  \centering
  \includegraphics[width=\linewidth]{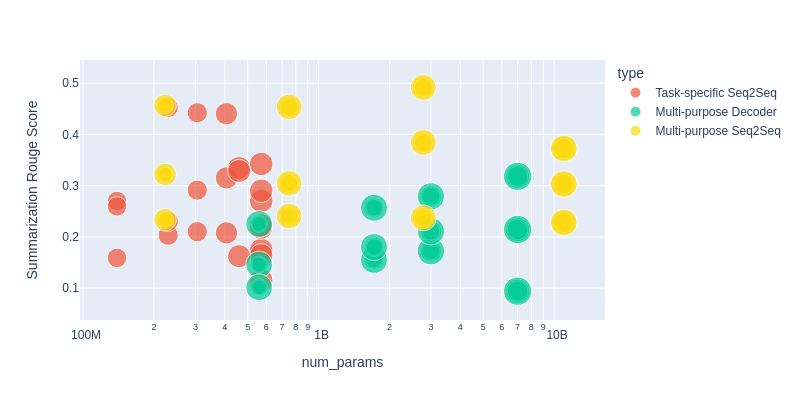}
  \caption{A plot of model size, measured in number of parameters (x axis, in logarithmic scale) and summarization accuracy (y axis), with dot size indicating the quantity of emissions.}
\end{figure}

\begin{figure}[h!]
  \centering
  \includegraphics[width=\linewidth]{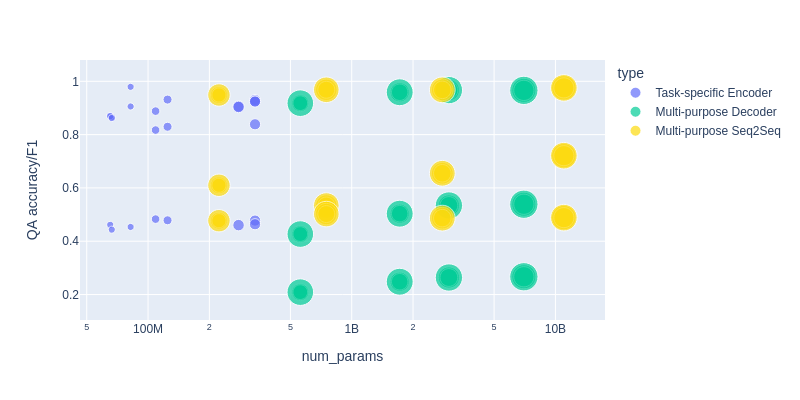}
  \caption{A plot of model size, measured in number of parameters (x axis, in logarithmic scale) and question answering accuracy (y axis), with dot size indicating the quantity of emissions.}
\end{figure}

\end{document}